\documentclass[10pt,twocolumn,letterpaper]{article}

\usepackage{cvpr}              

\usepackage{graphicx}
\usepackage{amsmath}
\usepackage{amssymb}
\usepackage{booktabs}

\usepackage[pagebackref,breaklinks,colorlinks]{hyperref}

\usepackage[capitalize]{cleveref}
\crefname{section}{Sec.}{Secs.}
\Crefname{section}{Section}{Sections}
\Crefname{table}{Table}{Tables}
\crefname{table}{Tab.}{Tabs.}

\def\cvprPaperID{*****} 
\def\confName{CVPR}
\def\confYear{2024}

\begin{document}

\title{Optimizing Split Points for Error-Resilient SplitFed Learning}
\author{Chamani Shiranthika, Parvaneh Saeedi, Ivan V. Baji\'{c} \\
School of Engineering Science, Simon Fraser University, Burnaby, BC, Canada.\\
{\tt\small\{csj5, psaeedi, ibajic\}@sfu.ca}
}
\maketitle

\begin{abstract}
Recent advancements in decentralized learning, such as Federated Learning (FL), Split Learning (SL), and Split Federated Learning (SplitFed), have expanded the potentials of machine learning. SplitFed aims to minimize the computational burden on individual clients in FL and parallelize SL while maintaining privacy. This study investigates the resilience of SplitFed to packet loss at model split points. It explores various parameter aggregation strategies of SplitFed by examining the impact of splitting the model at different points—either shallow split or deep split—on the final global model performance. The experiments, conducted on a human embryo image segmentation task, reveal a statistically significant advantage of a deeper split point.
\end{abstract}

\section{Introduction}
\label{sec:intro}
Federated learning (FL)~\cite{mcmahan_2017} allows multiple clients to train machine learning models without sharing data, which is particularly advantageous for privacy-sensitive domains like healthcare. However, FL requires all clients to train models locally, causing challenges for resource-constrained clients. Split Learning (SL)~\cite{Gupta_2018,Vepakomma_2018} addresses this by splitting the model between clients and a server. Split Federated Learning (SplitFed) ~\cite{Thapa_2022,shiranthika_2023} combines FL's privacy preservation with SL's model balancing, offering the best of both.

Error resilience is crucial in decentralized learning. Recent research has explored SplitFed's robustness to annotation errors~\cite{Zahra_isbi_2023} and noisy communication links \cite{Zahra_icassp_2023}, but packet loss—a common transmission error—has not paid attention yet. Prior work in FL has tackled packet loss by modelling communication links as packet erasure channels~\cite{shirvani_2022} or implementing loss-tolerant strategies \cite{zhou2021loss}. In SL, packet loss occurs at model split points, challenging the optimal split points selection for loss resilience. Research on optimal split points selection in SL\cite{xu2023accelerating} and in collaborative intelligence exists \cite{kim_splitnet,tuli_splitplace_2022}. Loss resilience studies in these domains are separate \cite{ALTeC_Access2020,Inpainting_ICC2021,CALTeC_ICIP2021, kang_neurosurgeon_2017,bajic2021collaborative, eshratifar2019bottlenet}. Analyzing split point choices based on loss resilience is not previously studied.

This is the first statistical analysis of the effect of model split points in SplitFed on loss resilience. We examine various parameter aggregation methods ($Param_{agg}$) under different conditions, such as the packet loss probabilities ($P_L$) and the number of clients experiencing packet loss ($N_c$).

\begin{figure}[t]
\centerline{\includegraphics[scale = 0.35]{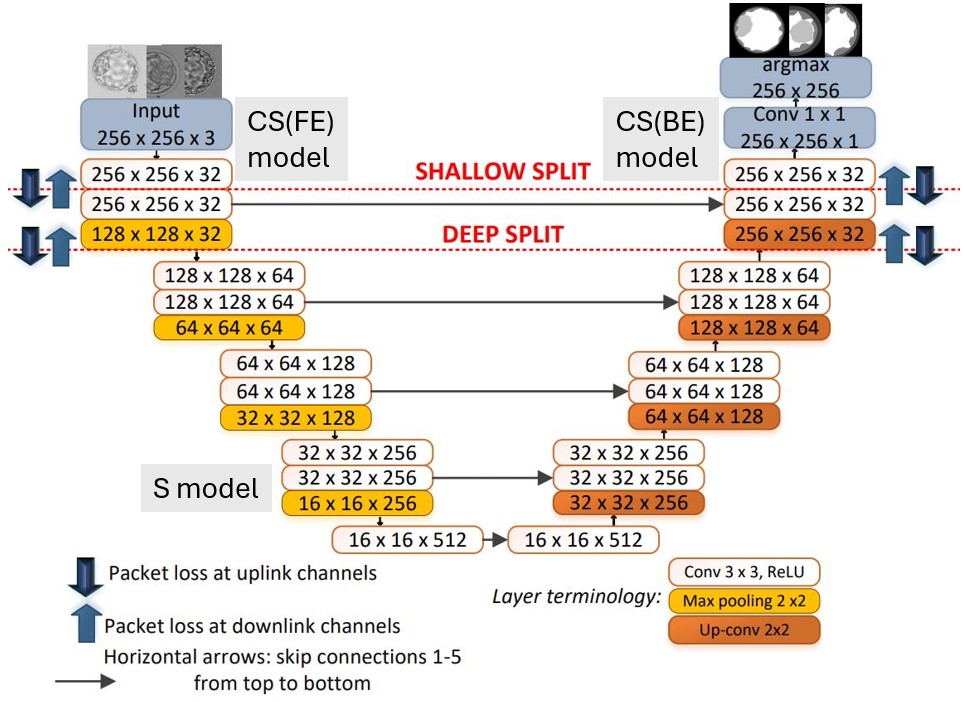}}
\caption{{\small{Split U-Net architecture}}}
\label{Split_model}
\end{figure}

\section{Methodology}
\label{sec:methodology}
We utilized a Split U-Net model for human embryo component segmentation, as in Fig.~\ref{Split_model}. Three split models were: client-side front-end (CS(FE)), server-side (S), and client-side back-end (CS(BE)). Two split points were: the shallow split and the deep split. We used the Blastocyst dataset \cite{lockhart_2019}, which has $781$ human embryo images with ground-truth segmentation masks for five components: background, zona pellucida (ZP), trophectoderm (TE), inner cell mass (ICM), and blastocoel (BL). $70$ images were assigned for testing. Data were non-uniformly distributed among five clients. Each client allocated $85\%$ of its data for training and $15\%$ for validation. During training, resizing, horizontal flipping, and vertical flipping were applied. The loss function was Soft Dice, and the optimizer was Adam with an initial learning rate of $1\times10^{-4}$. The Jaccard index without background was employed as the performance metric. The system was trained for $12$ local and $15$ global epochs. Each experiment was repeated for $10$ runs and the mean JI (MJI) was recorded. Packet loss was simulated, with each lost packet representing a zeroed-out row of feature and gradient maps. $P_L\in\{0.1, 0.3, 0.5, 0.7, 0.9\}$, and was independent and identically distributed (iid). $N_c\in\{0, 1, 2, 3, 4, 5\}$. $Param_{agg}\in\{$naive averaging~\cite{mcmahan_2017}, federated averaging (FedAvg)~\cite{mcmahan_2017}, auto-FedAvg~\cite{Xia_2021}, fed-NCL\_V2~\cite{Li_2021}, fed-NCL\_V4~\cite{Li_2021}\}.  

\section{Experimental results}
\label{sec:experiments}
\subsection{Experiments without packet loss}
Our U-Net model, centrally trained on the same dataset\cite{Saeedi_2017} as BLAST-NET 
(the sole network accessible for human embryo component segmentation)
~\cite{Rad_2019}, achieved a MJI of $81.70\%$, compared to BLAST-NET's MJI, which was $79.88\%$. This shows our model's favourable  performance compared to BLAST-NET. The SplitFed U-Net model's MJI for naïve averaging, FedAvg, auto-FedAvg, fed-NCL\_V2, and fed-NCL\_V4 were $82.78\%$, $82.57\%$, $82.99\%$, $83.02\%$, and $82.95\%$, respectively. We performed $20$ ($\binom{5}{2} \times 2$) pairwise t-tests for the difference in these MJIs. $J_{\text{param\_agg1}}$ and $J_{\text{param\_agg2}}$ are MJIs of two of $Param_{agg}$. The two-tailed t-test is
\begin{equation}
    \begin{aligned}
        H_0: J_{\text{param\_agg1}} = J_{\text{param\_agg2}} \\
        H_1: J_{\text{param\_agg1}} \neq J_{\text{param\_agg2}}  
    \end{aligned}
\end{equation}
In most of the cases, $p$-value~\cite{fisher_1936} $< 0.05$, so the null hypothesis $H_0$ can be rejected and we can conclude that the $param_{agg}$ are statistically different. 

\subsection{Experiments with packet loss}
\begin{figure}
\centerline{\includegraphics[scale=0.60]{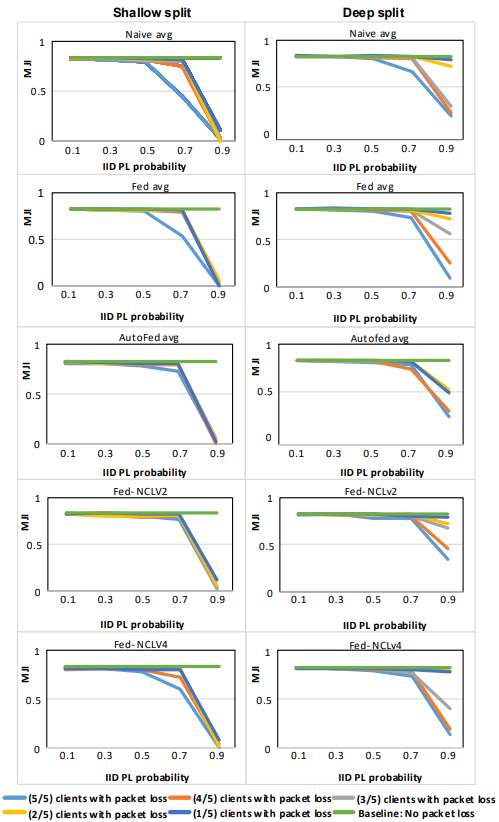}}
\caption{ {\small MJI vs. $P_L$ for shallow split (left) and deep split (right), with variying $N_c$ in the SplitFed U-Net model.} }
\label{deepvsshallow}
\vspace{-10pt}
\end{figure}

Then we trained our SplitFed U-Net model with the two splits across all $P_L$. The deep split model outperformed the shallow split model, as in Fig.~\ref{deepvsshallow}. To statistically confirm this, we conducted $125$ ($5 \times 5 \times 5$) pairwise t-tests for each combination of $P_L$ and $N_c$. $J_{\text{D}}$ and $J_{\text{S}}$ are MJIs of deep and shallow splits, respectively. The one-tailed t-test is:  
\begin{equation}
    \begin{aligned}
        H_0: J_{\text{D}} \leq J_{\text{S}} \\
        H_1: J_{\text{D}} > J_{\text{S}} \\
    \end{aligned}
\end{equation}
In all cases, $p$-value $< 0.05$, so the null-hypothesis $H_0$ can be rejected and we can conclude that deep split produces a higher MJI than shallow split. 
Finally, we analyzed whether certain $Param_{agg}$ perform better than others in the presence of packet loss for the deep split model. Conducting $250$ ($\binom{5}{2} \times 5 \times 5$) pairwise comparisons for each combination of $param_{agg}$, $P_L$, and $N_c$, we found performance variations among methods, but no consistent pattern emerged to prove one method's superiority over others.

\section{Conclusions}
In this study, we investigated the effect of model split points in SplitFed when clients experience packet loss. Our experiments, utilizing five advanced parameter aggregation methods, revealed a statistically significant advantage of a deeper split. Two reasons contribute to this: Firstly, the deep split model provides the CS(BE) with additional layers to recover the lost data. Secondly, in our deep split U-Net model, the initial skip connection is entirely at the client, enabling features and gradients transfers without subjecting to packet loss. Our SplitFed U-Net model showed resilience to packet loss, up to $50\%$, in both shallow and deep splits. This resilience may be due to the utilization of ReLU activation \cite{ALTeC_Access2020}, and the phenomenon of packet loss acting similarly to dropout. No significant performance gap was observed between models trained with packet loss probabilities up to 0.5, but those trained with higher packet loss probabilities exhibited diminished performance. More research would focus on exploring multiple SplitFed networks, investigating realistic packet loss models, and developing robust aggregation and recovery methods. 
This study will initiate further exploration of packet loss effects on SplitFed.

{\small
\bibliographystyle{ieee_fullname}
\bibliography{refs}
}

\end{document}